\pdfoutput=1
\documentclass[11pt]{article}
\usepackage[]{acl}

\usepackage{times}  %
\usepackage{helvet}  %
\usepackage{courier}  %
\usepackage{graphicx} %
\usepackage{natbib}  %
\usepackage{caption} %
\usepackage{algorithm}
\usepackage{algorithmic}

\usepackage{newfloat}
\usepackage{listings}
\DeclareCaptionStyle{ruled}{labelfont=normalfont,labelsep=colon,strut=off} %
\floatstyle{ruled}
\newfloat{listing}{tb}{lst}{}
\floatname{listing}{Listing}

\usepackage{enumerate}
\usepackage{microtype}
\usepackage{amsthm}
\usepackage{amsmath}
\usepackage{amsfonts,bm,amssymb}
\usepackage{soul,color}
\usepackage{enumitem}
\usepackage{titlesec}
\usepackage{graphicx}
\usepackage{multirow,multicol,makecell,booktabs}
\usepackage{array}
\usepackage{booktabs}
\usepackage{caption}
\usepackage{subcaption}
\usepackage{tipa}
\usepackage{stackengine}
\usepackage{rotating}
\usepackage{tabularx}
\usepackage{enumerate}
\usepackage{cleveref}
\usepackage{xcolor,pifont,bbding}
\usepackage[ruled,vlined,algo2e]{algorithm2e}

\crefname{algocf}{Alg.}{Algs.}
\Crefname{algocf}{Algorithm}{Algorithms}

\crefformat{section}{\S#2#1#3} %
\crefformat{subsection}{\S#2#1#3}
\crefformat{subsubsection}{\S#2#1#3}

\theoremstyle{definition}

\titlespacing{\paragraph}{0pt}{2ex}{0.2cm}

\newcommand\code[1]{\texttt{#1}}

\newcommand{\PreserveBackslash}[1]{\let\temp=\\#1\let\\=\temp}
\newcolumntype{C}[1]{>{\PreserveBackslash\centering}p{#1}}
\newcolumntype{R}[1]{>{\PreserveBackslash\raggedleft}p{#1}}
\newcolumntype{L}[1]{>{\PreserveBackslash\raggedright}p{#1}}

\newcommand*\colourcheck[1]{%
  \expandafter\newcommand\csname #1check\endcsname{\textcolor{#1}{\ding{52}}}%
}
\newcommand*\colourcross[1]{%
  \expandafter\newcommand\csname #1cross\endcsname{\textcolor{#1}{\ding{54}}}%
}

\definecolor{pastelgreen}{rgb}{0.16, 0.67, 0.53}

\colourcheck{green}
\colourcheck{pastelgreen}
\colourcross{red}

\title{Generating Intermediate Steps for NLI with Next-Step Supervision}
\date{}

\author{Deepanway Ghosal$^1$, Somak Aditya$^2$, Monojit Choudhury$^3$ \\\\
$^1$ ISTD, Singapore University of Technology and Design\\
$^2$ Department of CSE, IIT Kharagpur\\
$^3$ Turing India, Microsoft\\
\texttt{deepanway\_ghosal@mymail.sutd.edu.sg}\\
\texttt{saditya@cse.iitkgp.ac.in}\\ 
\texttt{monojitc@microsoft.com}
}

\begin{document}
\maketitle

\begin{abstract}
The Natural Language Inference (NLI) task often requires reasoning over multiple steps to reach the conclusion. While the necessity of generating such intermediate steps (instead of a summary explanation) has gained popular support, \textit{it is unclear how to generate such steps without complete end-to-end supervision and how such generated steps can be further utilized}. In this work, we train a sequence-to-sequence model to generate only the next step given an NLI premise and hypothesis pair (and previous steps); then enhance it with external knowledge and symbolic search to generate intermediate steps with only next-step supervision. We show the correctness of such generated steps through automated and human verification. Furthermore, we show that such generated steps can help improve end-to-end NLI task performance using simple data augmentation strategies, across multiple public NLI datasets. 

\end{abstract}

\section{Introduction}
\begin{table*}[ht!]
\small
\centering
\begingroup
\resizebox{0.95\linewidth}{!}
{\begin{tabular}{L{5.2cm}L{3.5cm}L{5.3cm}cccc}
\toprule
\multirow{2}{*}{\textbf{Premise (P)}} & \multirow{2}{*}{\textbf{Hypothesis (H)}} & \multirow{2}{*}{\textbf{Proof}} & \multicolumn{2}{c}{\textbf{Correctness}} & \multicolumn{2}{c}{\textbf{Minimality}}\\
& & & C1 & C2 & M1 & M2 \\
\midrule
\multirow{3}{5.2cm}{The wind propels a sailing ship on a group of cruisers.} & \multirow{3}{3.5cm}{There are many boats out.} & \textbf{P $\rightarrow$ I1:} A group of cruisers are in the water. \textbf{$\rightarrow$ I2:} A group of boats are in the water. \textbf{$\rightarrow$ H} & \hfil\multirow{3}{*}{\pastelgreencheck} & \hfil\multirow{3}{*}{\pastelgreencheck} & \hfil\multirow{3}{*}{\pastelgreencheck} & \hfil\multirow{3}{*}{\pastelgreencheck}\\

\midrule
A young girl wearing a red shirt and cap smiling and holding a small toy is standing in front of a group of children playing behind her. & \multirow{4}{3.5cm}{The young girl is wearing a red shirt.} & \multirow{4}{5.3cm}{\textbf{P $\rightarrow$ I1:} A young girl in a red shirt and cap is in front of others. \textbf{$\rightarrow$ I2:} The young girl is wearing a cap and red shirt. \textbf{$\rightarrow$ H}} & \hfil\multirow{3}{*}{\pastelgreencheck} & \hfil\multirow{3}{*}{\pastelgreencheck} & \hfil\multirow{3}{*}{\redcross} & \hfil\multirow{3}{*}{\redcross}\\

\midrule
\multirow{3}{5.2cm}{Numerous customers browsing for products in a market.} & \multirow{3}{3.5cm}{People are shopping.} & \textbf{P $\rightarrow$ I1:} A group of people are shopping. \textbf{$\rightarrow$ I2:} There are a bunch of people shopping. \textbf{$\rightarrow$ H} & \hfil\multirow{3}{*}{\pastelgreencheck} & \hfil\multirow{3}{*}{\pastelgreencheck} & \hfil\multirow{3}{*}{\redcross} & \hfil\multirow{3}{*}{\pastelgreencheck} \\

\midrule
A gentleman with his eyes closed playing an old fluglehorn into a microphone. & \multirow{2}{3.5cm}{A man plays an instruement with his eyes closed.} & \textbf{P $\rightarrow$ I1:} The gentleman is listening to music. \textbf{$\rightarrow$ H} & \hfil\multirow{2}{*}{\pastelgreencheck} & \hfil\multirow{2}{*}{\redcross} & \hfil\multirow{2}{*}{\redcross} & \hfil\multirow{2}{*}{\pastelgreencheck} \\
\bottomrule
\end{tabular}
}
\endgroup
\caption{Correctness and minimality of proofs as outlined in \Cref{sec:proof-definition}.}
\label{tab:main-examples}
\end{table*}

Complex NLP tasks such as Natural Language Inference (NLI) and Question-Answering (QA) often requires reasoning over multiple steps using multiple facts and implicit commonsense knowledge ~\cite{trivedi-etal-2020-multihop,DBLP:conf/aaai/SapBABLRRSC19,camburu2018snli}. It has been long argued that \cite{DBLP:journals/cacm/Lipton18}, the state-of-the-art Deep Learning models should also output some sort of explanation (such as intermediate steps or a textual explanation) alongwith the final answer. The opaque performance and poor out-of-distribution generalization performance of Transformers-based models  \cite{kaushik2019learning,ribeiro-etal-2020-beyond} have refuelled this discussion. However, it is unclear how these intermediate steps can be generated for unconstrained natural language Premise-hypothesis pairs (such as in crowd-sourced NLI datasets) as it is non-trivial to collect crowd-sourced fine-grained explanations or generate them synthetically. Most importantly, it is unclear how such steps can be utilized further for the NLI task.

Despite some efforts \cite{camburu2018snli} crowd-sourced collection of intermediate steps comes with complications, as human-written explanation can be subjective, and it is hard to automatically verify or utilize such explanations. Recently, researchers have explored synthetic generation of intermediate steps (or \textit{proof} trees) \cite{DBLP:conf/ijcai/ClarkTR20,DBLP:conf/acl/TafjordDC21,DBLP:conf/emnlp/SahaGSB20}, where the main goal was to test whether Transformers (the backbone for NLP models) can perform deductive reasoning over natural language statements. Provided their examples come from an underlying symbolic system with closed world rules and facts, it is unclear how this strategy of generating proof tree can be extended to unconstrained natural language premise-hypothesis pairs. In comparably structured domains, such as code snippet retrieval, knowledge-graph based QA and symbolic mathematics, \citet{DBLP:journals/corr/abs-2112-00114,DBLP:conf/wsdm/HeL0ZW21,agarwal2021analyzing} have shown how to generate such intermediate steps automatically, and how utilizing such steps can enhance end-to-end task performance. 

We take inspiration from these domains. We assume that an ideal NLI system reaches a series of intermediate conclusions to derive the final conclusion (entailment/contradiction/neutral) and we call these intermediate steps as \textit{proof}. Since such \textit{proof}s can widely vary linguistically, logically and in length; to make verification and generation easier, we impose various constraints of \textit{correctness}, \textit{minimality}, and \textit{atomicity} on what we expect as natural language \textit{proof}s for a given premise-hypothesis pair. Alongwith human verification, we propose corresponding automated metrics for verification. 
As human annotation and synthetic generation of such groundtruth proofs is non-trivial, we utilize single-step supervision to train a T5 encoder-decoder model \cite{raffel2019exploring} on various available entailment datasets; learning various aspects of proof generation from SNLI, Monotonicity Entailment Dataset, and Entailment Bank. To enrich single-step generation with external knowledge, we build a fact retriever and sentence composition model; that retrieves facts from external knowledge bases and learns to deduce new facts. We then explore search-based methods that utilizes the fact-augmented T5 model to generate multiple-step \textit{proof}s. Our automated and human verification results show the efficacy of the proof generation process. We further use these generated proofs as additional training data and show improvement in end-to-end NLI task performance.  
 
Our \textbf{contributions} include, 1) using only next-step supervision to train a T5 encoder-decoder based model, and external fact composition model; and then use search to generate sequence of knowledge-enriched intermediate steps. We 2) propose and use automated and human verification metrics to show the correctness of the generated proofs. Lastly, 3) we also show that augmenting such proofs during training can help enhance an NLI system's performance on MNLI, and MED (under low-training data regimes). 

\section{Constrained Intermediate Steps or Proofs in the NLI Context}
\label{sec:proof-definition}
In NLI, a premise-hypothesis pair is provided in natural language, and the task is to determine whether the hypothesis is entailed by, contradicts, or is neutral with respect to the premise. Ideally, an NLI system follows a logical sequence of steps (or \textit{proof}) to come to such a conclusion. However, many valid proofs may exist due to linguistic variability and multiple ways of reaching a conclusion from the context. Hence, to make generation and verification easier, we resort to defining logical properties that valid \textit{proofs} should have.  %

Given a premise ($P$), a hypothesis ($H$), and an implicit knowledge base $KB$ \footnote{Following \cite{dagan2005pascal}, we assume $KB$ encodes commonly assumed knowledge as facts and rules.}, an NLI \textit{proof} $\langle P, H \rangle$ (or $\langle P, \neg H \rangle$ for contradiction) is a sequence of sentences ($Y_1,\ldots, Y_m$), where $Y_j$ is either an inferred intermediate step (denoted by $I_j$), or an external fact or rule (denoted by $F_j$). A valid \textit{proof} should satisfy the following properties: i) (Correctness 1) each step is either a generic rule, or a fact which is a entailed by the premise; ii) (Correctness 2) hypothesis (or negated hypothesis) is a valid entailment of premise and intermediate steps; iii) (Minimality 1) each proof upto $Y_j$ (including $H$) should be minimal proof for $\langle P, Y_j \rangle$, iv) (Minimality 2) the sentences in the intermediate step ($I_j$) should not be trivially decomposable (using common linguistic or logical constructs)\footnote{M1 captures redundancy (see E4 in Tab.~\ref{tab:main-examples}). M2 captures some aspects of \textit{atomicity}. Sentences should not be compound in nature. We prefer ``John is going to Paris. Mia is going to Paris.'' over ``John and Mia are going to Paris''.  It also entails that intermediate conclusions should correspond to a semantic frame (which can not be trivially decomposable without losing context). However the latter is quiet hard to verify.} and consecutive inferred steps should be sufficiently different, and v) (Order) a step can only be generated with the help of previous steps.

\begin{figure*}[t]
    \centering
        \includegraphics[width=0.9\linewidth]{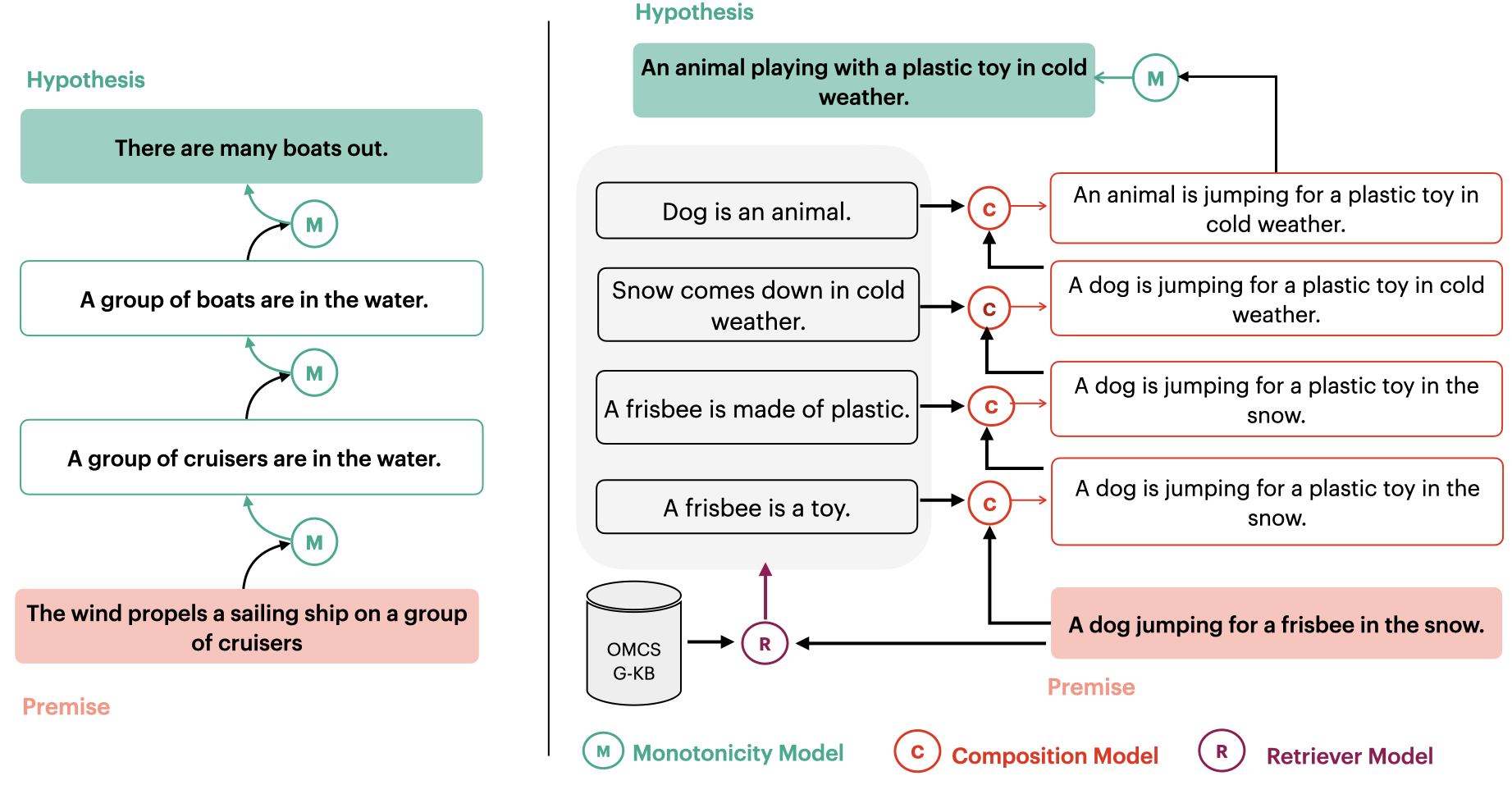}
        \caption{We illustrate two types of proofs that we explore: a) inference chains, and b) proofs with external facts. These proofs have been generated using \textit{unconstrained proof search} (\Cref{sec:ups}) and \textit{proof search with facts} (\Cref{sec:psf}).}
        \label{fig:model}
\end{figure*}

We provide few example proofs in Table~\ref{tab:main-examples}, highlighting the properties of correctness and minimality. These constraints motivate our method to generate \textit{proof}s for arbitrary P-H pairs with only next-step supervision. In Figure \ref{fig:model}, we show two types of proof that we generate, i) inference chains (i.e. only intermediate entailments) and ii) proofs with external (commonsense) facts (or rules) in sequence. For neutral cases, such properties are hard to define. Inspired from \citet{DBLP:conf/acl/KumarT20}, we proposed a method for intermediate step generation of neutral instances which is used only for data augmentation experiments.

\section{Provers: Generating Proofs with Next-Step Supervision }

A typical possibility to build provers is by training on an NLI dataset with groundtruth proofs, collected using crowd-sourcing or created using formal logic. However, crowd-sourced explanations are highly subjective and diverse; hence not easily verifiable. Similarly, following ProofWriter, generating formal logic-based proof is possible but hard to scale for arbitrary \textit{P-H} pairs. We instead rely on various NLI datasets, from which models can learn multiple ways of generating entailments, contradictions, and intermediate steps. We then enhance such provers with facts retrieved from knowledge bases and sentence composition methods. We use these techniques effectively to search for more generalized multiple-step proofs. We start from the premise \textit{P} and generate the steps of the proof recursively with the \textit{Prover} model. We use the term \textit{Prover} to collectively denote the models in \Cref{sec:model-prover} and \Cref{sec:model-fact-compose} as discussed next.

\subsection{Multi-Task Supervision}
\label{sec:model-prover}
We train a single T5-Large model with various objectives to generate inferences
from the premise. The objectives are specified by prefix tokens in the input text. The objectives are as follows.

\vspace{1mm}
\noindent \textbf{Entailed Sentence Generation}: The model is trained to generate possible entailments $E$ from the premise $P$. The entailment instances of the SNLI dataset~\cite{bowman-etal-2015-large} is used for training. The hypothesis $H$ is considered as $E$ during training. The input to the model is \code{entail: P} and the output to be generated is \code{H}.

\vspace{1mm}
\noindent \textbf{Contradictory and Neutral Sentence Generation}: The model is trained to generate possible contradictions and neutral inferences from the premise $P$. We use the relevant instances in the SNLI dataset for this objective. The input to the model is \code{contradict: P} or \code{neutral: P} and the output to be generated is the corresponding contradictory or neutral hypothesis \code{H}.

\vspace{1mm}
\noindent \textbf{Monotonic Sentence Generation:} The model is trained to generate monotonic inferences $M$ from the premise $P$. The input is \code{monotonic: P} and the output to be generated is \code{M}. We use the Monotonicity Entailment Dataset (MED)~\cite{yanaka-etal-2019-neural} for this objective. 
\vspace{1mm}

We merge and shuffle instances from the respective datasets to ensure that training is performed for all the objective functions simultaneously. More details can be found in the Appendix.

\subsection{Fact Retriever and Sentence Composition}
\label{sec:model-fact-compose}

\paragraph{Fact Retriever} The generator model (\cref{sec:model-prover}) can not generate proofs, %
which need reference to some external facts, or commonsense knowledge. Such knowledge may not be readily available in the generator model. It is thus necessary to assist the proof generation algorithm with relevant factual knowledge
to generate accurate and complete proofs.

We use sentences in Open Mind Commonsense (OMCS)~\cite{singh2002open} and GenericsKB~\cite{bhakthavatsalam2020genericskb} as the knowledge base (KB). The sentence embedding model \textit{all-mpnet-base-v2}
~\cite{reimers-2019-sentence-bert}
is used to retrieve facts for a given \textit{(premise, hypothesis)} pair from the KB. 
For a sentence $s$, we retrieve the facts $F$ from KB based on highest embedding cosine similarity in the following way:

\begin{itemize}[leftmargin=*, topsep=0.2pt]
\setlength\itemsep{-0.2em}
    \item \textit{Retrieval with key-words:} The list of noun tokens -- $n_p, n_h$ are extracted from the premise and the hypothesis. The token list is then joined together to form $s$ to perform retrieval.
    \item \textit{Retrieval with clustered key-words:} $n_p, n_h$ are divided into small groups of related words using clustering with word embeddings. For instance, 
    the following groups are created for the example in \Cref{fig:model}: \{dog, animal\}, \{snow, cold\}, \{frisbee, toy, plastic\}. Each of the groups is then merged together in a single string $s$ for performing retrieval.
\end{itemize}

\paragraph{Sentence Composition} We train another T5-Large model to generate compositions from a pair of input sentences. The model is trained on sentence triplets from the Entailment Bank~\cite{dalvi2021explaining} and RuleTaker~\cite{DBLP:conf/ijcai/ClarkTR20} datasets. Few example triplets $\langle S_1, S_2, S_3\rangle$ are as follows: $\langle$\textrm{Bob is green}, \textrm{All green people are rough}, \textrm{Bob is rough}$\rangle$, and $\langle$\textrm{Eruptions produce ash clouds}, \textrm{Ash blocks sunlight}, \textrm{Eruptions block sunlight}$\rangle$. Here, $S_1$ and $S_2$ can be composed to conclude $S_3$. In this setup, the input to the model is \code{conclude: $S_1$ <sep> $S_2$} and the output to be generated is \code{$S_3$}. The sentence composition model is then used with the fact retriever model to generate the proofs (\cref{sec:psf}).

\begin{table*}[ht!]
\small
\centering
\begingroup
\resizebox{0.94\linewidth}{!}
{\begin{tabular}{L{2.6cm}L{4.8cm}L{3.6cm}L{6.4cm}}
\toprule
\textbf{Method} & \textbf{Premise (P)} & \textbf{Hypothesis (H)} & \textbf{Proof}\\
\midrule
\begin{tabular}{@{}L{2.6cm}} \textbf{Unconstrained Proof Search} \end{tabular} 
&  
\begin{tabular}{@{}L{4.8cm}} An old woman in a white hat and purple and blue clothes is sitting down by a wooden building. \\ \end{tabular}
&  
There is a building.
& 
\begin{tabular}{@{}L{6.4cm}}
\textbf{P $\rightarrow$ I1:} A woman is sitting by a wooden building.  \\
\textbf{I1 $\rightarrow$ I2:} The building is made out of wood. \\
\textbf{I2 $\rightarrow$  H}
\end{tabular}
\\
\midrule
\multirow{5}{2.6cm}{\begin{tabular}{@{}L{2.6cm}} \textbf{Fact Composition}  \end{tabular}}  & 
\begin{tabular}{@{}L{4.8cm}} Bicyclist ride the course near the ocean as the day comes to an end. \end{tabular}
& 
\begin{tabular}{@{}L{3.6cm}} The cyclist was riding near the ocean at sunset. \end{tabular}
& 
\begin{tabular}{@{}L{6.4cm}} 
\textbf{F1:} Sunsets can happen at the end of the day. \\ 
\textbf{F1 $\&$ P $\rightarrow$ I1:} A cyclist ride the course near the ocean during the sunset. \\
\textbf{I1 $\rightarrow$ H}
\end{tabular}
\\
\cline{2-4}\noalign{\vskip 1mm} 
 &
\begin{tabular}{@{}L{4.8cm}} A baby girl and little boy are standing next to a guitar and a drum. \end{tabular}
& 
\begin{tabular}{@{}L{3.6cm}} The girl is near an instrument. \end{tabular}
&
\begin{tabular}{@{}L{6.4cm}} 
\textbf{F1:} A drum is a percussion instrument. \\ 
\textbf{F1 $\&$ P $\rightarrow$ I1:} A baby girl and little boy are standing next to a percussion instrument. \\
\textbf{I1 $\rightarrow$ I2:} A baby girl standing next to a percussion instrument. \\
\textbf{I2 $\rightarrow$ H}
\end{tabular}
\\
\bottomrule
\end{tabular}
}
\endgroup
\caption{Examples of generated proofs using different methods.}
\label{tab:examples}
\end{table*}

\section{Proof Generation}
We generate proofs with two different methods as described below, and provide some examples of generated proofs in \Cref{fig:model} and \Cref{tab:examples}.

\subsection{Unconstrained Proof Search (Inference Chains)}
\label{sec:ups}
The \textit{entailment} and \textit{monotonic sentence} generation setup use the premise $P$ to create possible implications/inference chains. We use this generation setup in a recursive manner with level search (or beam search) to find multistep proofs $(I_1, I_2)$ from the collection of generated chains. The recursive generation and search is performed as follows.

\begin{itemize}[leftmargin=*, topsep=0.2pt]
\setlength\itemsep{-0.2em}
    \item 
    Given $P$, we first generate one-step implications $\hat{i_1}$ from the T5 generator.
    \item 
    We denote the $n$ closest implications to the hypothesis $H$ as the filtered set $i_1$. The closeness measure is performed by using highest cosine similarities with $H$ using the sentence embedding model \textit{all-mpnet-base-v2}~\cite{reimers-2019-sentence-bert}.
    \item 
    $i_1$ is then to generate the next set of implications $\hat{i_2}$, which are filtered again to obtain $i_2$. 
    \item 
    The $n$ implications in $i_2$ and their respective source sentences in $i_1$ form the multistep proof set $(I_1, I_2)$ for level search.
    \item 
    The top $n$ implications (according to closeness with $H$) from the merged set of $i_1$ and $(i_1, i_2)$ form the proof set 
    for beam search.
\end{itemize}

We use the terminology \textit{unconstrained with similarity} 
for the above proof generation algorithm with sentence embedding based closeness measure. We use $n=10$ in our experiments. The generation and filtering process can be performed repeatedly to form proofs with more steps: $(I_1, I_2, .. I_m)$. However, we observe diminishing results after $I_2$, as steps tend to become repetitions of each other. 

\subsection{Proof Search with External Facts}
\label{sec:psf}

Given a premise, hypothesis pair $(P, H)$, we denote the facts retrieved from $KB$ as $F$. Let $F$ consist of $m$ distinct facts $\{F_1, F_2, .., F_m \}$. 
We now use the sentence composition model as follows to generate the proof.

\begin{itemize}[leftmargin=*, topsep=0.2pt]
\setlength\itemsep{-0.2em}
    \item We compose premise $P$ with each of the facts $\{F_1, F_2, .., F_m \}$ to generate the implications $\{I_1, I_2, .., I_m \}$. 
    \item If any composition $I_k$ is farther from $H$ than $P$, then the respective fact $F_k$ is discarded. The distance is computed using embedding cosine similarity from \textit{all-mpnet-base-v2}. We denote the filtered fact set as $\bar{F}$. 
    \item Facts from $\bar{F}$ are first ordered based on their ranking from the retriever. This ordered set $\{\bar{F}_1, \bar{F}_2, .., \bar{F}_j \}$ is composed with $P$ and the respective outputs recursively: 
    $P \ \&\ \bar{F}_1 \rightarrow \bar{I}_1$, followed by
    $\bar{I}_1 \ \&\ \bar{F}_2 \rightarrow \bar{I}_2$, ... followed by 
    $\bar{I}_{j-1} \ \&\ \bar{F}_j \rightarrow \bar{I}_j$. %
    \item If $\bar{I}_j$ is not sufficiently close to $H$, then further implications $\hat{I}_j$ are generated using the \textit{entailment} and \textit{monotonic sentence} generator model.
\end{itemize}

\section{Experimental Results}
We perform two sets of experiments -- i) we use human and automatic evaluation to evaluate different aspects of the proofs generated from our proposed method, and ii) we use the generated proofs as additional labeled data for NLI and analyze its effect on the NLI classification task performance.

\begin{table*}[ht!]
\centering
\resizebox{\linewidth}{!}{
	\begin{tabular}{l|c|c|c|c|cc|cc|cc|cc|c|ccc}
	\toprule
	\multirow{3}{*}{\textbf{Algorithm}} & \multicolumn{4}{c|}{\textbf{Correctness}} & \multicolumn{12}{c}{\textbf{Minimality}}\\
	& \multirow{2}{*}{$P-H$} & \multirow{2}{*}{$P-I_1$} & \multirow{2}{*}{$I_1-I_n$} & \multirow{2}{*}{$I_n-H$} & \multicolumn{2}{c|}{$P-H$} & \multicolumn{2}{c|}{$P-I_1$} & \multicolumn{2}{c|}{$I_1-I_n$} & \multicolumn{2}{c|}{$I_n-H$} & \multirow{2}{*}{$\#$ Steps} & \multicolumn{3}{c}{$\#$ Keywords} \\
	& & & & & B4 & JS & B4 & JS & B4 & JS & B4 & JS & & P & $I_1-I_n$ & H\\
	\midrule
	Gold Proofs & \multirow{3}{*}{95.27} & 94.59 & 98.42 & 94.09 & \multirow{3}{*}{0.0272} & \multirow{3}{*}{66.26} & 0.0981 & 68.34 & 0.2223 & 74.07 & 0.0911 & 65.86 & \multirow{3}{*}{2} & \multirow{3}{*}{6.82} & 2.94 & \multirow{3}{*}{2.54} \\
	Negated Gold Proofs & & 3.94 & 9.45 & 20.08 & & & 0.0189 & 61.61 & 0.0596 & 62.59 & 0.0502 & 59.91 & & & 2.83 & \\
	Perturbed Gold Proofs & & 5.91 & 3.15 & 24.41 & & & 0.0139 & 62.37 & 0.0202 & 58.30 & 0.0147 & 56.80 & & & 3.10 & \\
	\midrule
	T5 w/o Search & 92.28 & 94.06 & 94.63 & 36.25 & 0.0751 & 71.07 & 0.0793 & 67.36 & 0.3070 & 78.82 & 0.1162 & 67.45 & 2 & 7.14 & 2.74 & 3.27 \\
	\midrule
	UC Level Search & \multirow{2}{*}{92.28} & 92.46 & 89.85 & 67.85 & \multirow{2}{*}{0.0751} & \multirow{2}{*}{71.07} & 0.1367 & 70.97 & 0.2443 & 75.94 & 0.0987 & 68.76 & 2 & \multirow{2}{*}{7.14} & 3.25 & \multirow{2}{*}{3.27} \\
	UC Beam Search &  & 91.99 & 82.55 & 72.94 &  &  & 0.1308 & 69.81 & 0.2652 & 75.82 & 0.2240 & 76.09 & 1.71 & & 3.21 & \\
	\midrule 
	Proof Search with Facts & 92.42 & 68.69 & 67.68 & 90.15 & 0.0066 & 65.52 & 0.6261 & 90.10 & 0.3540 & 80.63 & 0.0826 & 71.05 & 2.93 & 7.36 & 4.44 & 2.92 \\
    \bottomrule
	\end{tabular}
	}
	\caption{Evaluation of proofs with correctness and minimality metrics for: i) gold, noisy gold proofs ii) unconstrained (UC) level, beam search in all SNLI test set entailment pairs, and iii) proof search with facts in a subset of SNLI test set entailment pairs. 
	$P, H, I_1, I_n$, B4, JS denotes the premise, hypothesis, first, last step of the proof, BLEU-4, and Jaccard similarity respectively. $I_1-I_n$ denotes the average score of all intermediate pairs. Correctness scores indicate the percentage of pairs predicted as entailment by the \textit{roberta-large-mnli} model. Minimality scores are computed with BLEU-4, Jaccard similarity, number of proof steps and number of keywords.}
	\label{tab:results-multi}
\end{table*}

\subsection{Human Verification of Proofs}
We perform human verification on a subset of generated proofs in the SNLI dataset.
We train four CS graduate students (trained in NLP) with explicit instructions (in Appendix). We select top-2 proofs from \textit{unconstrained proof search} for 500 randomly selected SNLI entailment instances -- resulting in a total of 1000 proofs. Human annotators score each proof based on their correctness and minimality. For a proof with two intermediate sentences, a correctness score of 3, 2, 1, or 0 is given, signifying all three steps are correct, the first two steps are correct, only the first step is correct, and all other cases, respectively. In this schema, steps at the beginning have more importance than a step towards the end. Annotators also give a minimality score of 1 or 0, signifying all steps are minimal or at least one step is non-minimal, respectively. The human verification process results in a normalized correctness score of 67.28\% and a minimality score of 0.7517. More details can be found in the appendix.

\subsection{Automatic Evaluation of Proofs}
\label{sec:auto-eval}
\paragraph{How to Evaluate the Generated Proofs?} We also automatically assess the quality of proofs generated using \textit{unconstrained proof search} and \textit{proof search with facts} methods.
We design a number of metrics corresponding to the constraints over the proofs as introduced in \Cref{sec:proof-definition}. 
For \textbf{Correctness}, we use the entailment probability score from \textit{RoBERTa-MNLI}. \textbf{Minimality} is more non-trivial to measure. We instead use proxy metrics: \textbf{i)} BLEU4 and Jaccard similarity to quantify distinctiveness of consecutive steps, \textbf{ii)} Avg. number of keywords (nouns, verbs and adjectives) to quantify atomicity of steps, and \textbf{iii)} Avg. number of steps for generated proofs.

Assume an entailment instance from SNLI: $(Pre, Hyp)$ whose proof is the sequence $(I_1, I_2)$. We compute the correctness and minimality for all consecutive step pairs $(Pre-I_1), (I_1-I_2),$ and $(I_2-Hyp)$. 
A high percentage of step pairs to be predicted as entailment is desired for correctness. For minimality (BLEU-4 and Jaccard), we expect scores to be neither very high nor very low. A very high score (close to 1) would suggest that the pair is not sufficiently different. A very low score (close to 0) would suggest that the pair may not be related at all. Similarly, we expect length of proofs to be not too high nor too low (such as zero or single-step proofs). To calibrate this range and set a baseline, we first report these metrics in \Cref{tab:results-multi} on a subset of gold proofs that are evaluated to be correct by human annotators. We also systematically introduce noise to observe how these metrics behave. %

\noindent \textbf{Automatic Evaluation of Gold Proofs:} The suitability of the above metrics can be questioned for two reasons. Firstly, we do not have any corresponding gold proofs (that satisfies the constraints in \Cref{sec:proof-definition}). %
Second, it is hard to guess the range (or the best value) for the metrics, as we use proxy metrics and neural network models that are prone to noise. To mitigate these issues, 
we select a set of 200 proofs from \textit{unconstrained proof search} in SNLI, judged as correct and minimal by human annotators. We call this the \textit{Gold Proofs} set and measure its correctness and minimality through the automatic metrics. The \textit{Gold Proofs} are correct from human judgement and are also estimated as highly accurate from the automated correctness metric in \Cref{tab:results-multi}. The BLEU4 and Jaccard similarity (for minimality) range between 0.09 to 0.23 and 65\% to 75\%, respectively. Together, they establish the range for the metrics that we expect valid proofs should exhibit. When gold proofs are negated or perturbed (more in Appendix), we observe significant drops in correctness scores, and deviation from the minimality range. We thus expect the proposed metrics to be suitable proxies to quantify the properties of proofs in \Cref{sec:proof-definition}.

\noindent \textbf{Results for Unconstrained Proof Search:} We report automatic evaluation results of proofs generated from \textit{unconstrained proof search} (using level and beam search) in \Cref{tab:results-multi}. %
We select top-2 proofs from \textit{unconstrained proof search} for all entailment pairs in SNLI and report the mean scores in \Cref{tab:results-multi}. Results suggest that the \textit{unconstrained search} algorithm is capable of generating proofs with high correctness and ideal minimality. The minimality scores (BLEU4, Jaccard similarity) of the \textit{level search} generated proofs are very close to that of the \textit{Gold Proofs}. A general trend in the correctness measure is that the score decreases as we move from left to right pairs. This is expected as the hypothesis is not directly used for proof generation in \textit{unconstrained search}, resulting in the final pair being less entailing compared to previous ones.

\noindent \textbf{Results for T5 w/o Search:}  We also report results for the T5 model (\Cref{sec:model-prover}) where we don't use any search until the last $I_n \rightarrow H$ step. We obtain significantly lower correctness for the final step and mostly out of range minimality scores with this method. The T5 w/o search method only generates entailed statements from the premise. Thus we obtain high correctness scores in the initial steps, as reported in \Cref{tab:results-multi}. However, without guiding the intermediate steps towards the hypothesis a complete proof cannot be generated. This is why the correctness score of the last step is abysmally low for T5 w/o search. The minimality scores of the intermediate steps $(I_1 - I_n, I_n - H)$ for the unconstrained beam / level search method are also closer to the gold proofs compared to T5 w/o search.

\begin{table*}[ht!]
\small
\centering
\resizebox{0.8\linewidth}{!}{
	\begin{tabular}{cc|cccc|ccc}
	\toprule
	\multicolumn{2}{c|}{\textbf{Trained On}} & \multicolumn{4}{c|}{\textbf{MNLI}} & \multicolumn{3}{c}{\textbf{MED}} \\
	\textbf{MNLI \%} & \textbf{Proofs \%} & \textbf{Entailment} & \textbf{Neutral} &\textbf{Contradict} & \textbf{Average} & \textbf{Entailment} & \textbf{Neutral} & \textbf{Average} \\
	\midrule
	1 & 0 & 83.86 & 75.84 & 85.49 & 81.84 & 44.84 & 35.15 & 40.02 \\
	1 & 1 & 85.05 & 77.16 & 83.98 & 82.19 & 49.98 & 35.72 & 42.89 \\
	\midrule 
	2 & 0 & 86.26 & 79.38 & 86.22 & 84.06 & 48.40 & 39.33 & 43.89 \\
	2 & 2 & 86.08 & 79.93 & 86.68 & 84.32 & 51.34 & 39.68 & 45.54 \\
	\midrule 
	10 & 0 & 87.65 & 80.87 & 88.47 & 85.76 & 48.81 & 35.61 & 42.24 \\
	5 & 5 & 89.14 & \textbf{83.89} & \textbf{89.96} & 87.74 & 53.81 & 34.22 & 44.07 \\
	10 & 10 & \textbf{89.72} & 83.56 & 89.89 & \textbf{87.80} & \textbf{54.76} & \textbf{39.82} & \textbf{47.32}\\
    \bottomrule
	\end{tabular}
	}
	\caption{F1 scores for MNLI (val-matched) and MED datasets with the RoBERTa-Large model. Numbers on the \textbf{MNLI \%} column indicate the percentage of the MNLI train set instance used for training. Numbers on the \textbf{Proofs \%} column indicates the equivalent number of instances from the \textit{Prover} generated data used for training. Hence, \textbf{Proofs \%} of 1 implies the number of \textit{Prover} generated instances is the same as 1\% of MNLI train instances. Scores are average of 3 runs.}
	\label{tab:results-mnli-med}
\end{table*}
\begin{table*}[t]
\small
\centering
\resizebox{0.75\linewidth}{!}{
	\begin{tabular}{l|ccc|ccc}
	\toprule
	\textbf{Data (MNLI \%, Proofs \%)} $\rightarrow$ & 1\%, 0\% & 2\%, 0\% & 1\%, 1\% & 10\%, 0\% & 5\%, 5\% & 10\%, 10\% \\
	\textbf{Category} $\downarrow$ & & & & & & \\
	\midrule 
	Boolean & 55.23 & 60.58 & \textbf{62.46} & \textbf{62.83} & 60.09 & 60.5 \\
    Causal & \textbf{89.19} & 82.72 & 85.42 & 87.86 & 90.51 & \textbf{93.13} \\
    Comparative & 49.05 & 48.68 & \textbf{49.66} & 57.33 & 55.90 & \textbf{59.12} \\
    Conditional & 36.19 & \textbf{49.05} & 35.43 & 41.00 & \textbf{44.51} & 40.30 \\
    Coreference & 51.08 & \textbf{52.62} & 51.91 & 50.00 & 50.00 & \textbf{53.82} \\
    Implicature & 17.09 & \textbf{19.86} & 15.55 & 22.91 & \textbf{23.07} & 22.61 \\
    Lexical & 92.44 & \textbf{96.43} & 95.62 & 93.99 & \textbf{97.41} & 96.15 \\
    Negation & \textbf{97.86} & 97.61 & 91.42 & 99.61 & 99.81 & \textbf{99.95} \\
    Numerical & 44.01 & 51.62 & \textbf{56.74} & 59.48 & \textbf{70.10} & 62.13 \\
    Presupposition & 78.85 & 79.22 & \textbf{81.98} & 87.53 & \textbf{93.24} & 91.24 \\
    Quantifier & \textbf{65.40} & 64.78 & 60.30 & 65.20 & 67.71 & \textbf{68.65} \\
    Relational & 65.74 & 75.49 & \textbf{78.44} & 89.84 & 90.12 & \textbf{94.19} \\
    Spatial & \textbf{50.92} & 46.86 & 50.56 & 50.26 & 44.95 & \textbf{57.60} \\
    Syntactic & \textbf{83.37} & 81.69 & 76.08 & 93.88 & 99.30 & \textbf{99.92} \\
    Taxonomic & \textbf{76.24} & 65.89 & 64.36 & 64.73 & \textbf{72.34} & 60.23 \\
    Temporal & 28.63 & 27.26 & \textbf{37.89} & 45.62 & 44.78 & \textbf{47.29} \\
    World & 93.82 & \textbf{98.98} & 97.92 & \textbf{99.75} & 99.50 & 99.24 \\
    \midrule
    \textbf{Average} & 63.23 & \textbf{64.67} & 64.22 & 68.93 & 70.78 & \textbf{70.94} \\
    \bottomrule
	\end{tabular}
	}
	\caption{Reasoning category wise F1 scores on the Lo-NLI dataset for the RoBERTa-Large model. Scores are average of 3 runs.}
	\label{tab:results-lonli}
\end{table*}

\noindent \textbf{Results for Proof Search with Facts:} We identify entailment instances in SNLI where the cosine similarity between premise and hypothesis is less than 0.65 with a low overlap between tokens. We use \textit{proof search with facts} to generate proofs for these instances. In \Cref{tab:results-multi}, the minimality scores suggest that generated steps become more minimal as we move towards the hypothesis, having progressively less difference with the ideal minimality scores (of \textit{Gold Proofs}). Recursive composition of facts result in reduced correctness scores, as pairs are not always entailed  because of the introduction of new knowledge\footnote{Aligns with observations from \citet{joshi2020taxinli,tarunesh2021trusting} that RoBERTa-Large-MNLI model often fails on NLI pairs requiring external knowledge.}. However, the final step $I_n$ reaches a high correctness score of 90.15\%.
Furthermore, the monotonic sentence generator setup can be used on $I_n$ to generate inferences which are even closer to $H$ with average cosine similarity of 0.9489. 
Thus, the \textit{proof search with facts} algorithm is suitable for generating proofs that requires external commonsense knowledge.

\subsection{Usefulness of the Prover for NLI Tasks}
\label{sec:usefulness}
As illustrated in \cref{sec:model-prover}, the \textit{Prover} model is trained on various objective functions to generate inferences from the premise. Here, we show that these inferences could be used to improve end-to-end NLI task accuracy. We consider the \textit{entailed}, \textit{contradictory}, \textit{neutral}, and \textit{monotonic sentence generation} setup of \cref{sec:model-prover} and apply it on the premises of the SNLI dataset to create corresponding inferences. For a particular premise $P$, the inferences generated from the above four setups are considered to have a label of entailment, contradiction, neutral, and entailment, respectively.

The inferences generated from the \textit{Prover} are considered as additional labeled data for supervised learning in NLI tasks. In particular, we train RoBERTa-Large classification models for the MNLI~\cite{williams2018broad} dataset in a low-data regime. The models are trained with either i) MNLI-only data or ii) a mix of MNLI data and \textit{Prover} generated data. We show that performance consistently improves when trained with additional \textit{Prover} generated data, signifying the efficacy of our method. We also evaluate the models trained on MNLI in two other datasets -- Monotonicity Entailment Dataset (MED) and LoNLI~\cite{tarunesh2021trusting}. The incorporation of \textit{Prover} generated data helps in improving performance across most settings in these two datasets.

\noindent \textbf{Results for MNLI and MED:} We report results across various combination of amounts of MNLI data and \textit{Prover} generated data in \Cref{tab:results-mnli-med}. In the very low data regime (1\% MNLI + 0\% Proofs \textit{vs.} 1\% MNLI + 1\% Proofs), the additional data from \textit{Prover} helps in improving performance across all the label categories in MNLI and MED. However, the performance of 1\% MNLI + 1\% Proofs setting is lesser compared to 2\% MNLI + 0\% Proofs, signifying that more MNLI training data helps in better generalization for very low data setting. The improvement in performance is more prominent in the 10\% MNLI + 0\% Proofs \textit{vs.} 5\% MNLI + 5\% Proofs case. An identical number of training instances are used in both settings, and thus the improvement in performance can be directly attributed to the \textit{Prover} generated data. We observe improvements in all but one metric with large gains in MNLI neutral and MED entailment for 5\% MNLI + 5\% Proofs.

\noindent \textbf{Results for LoNLI:} The LoNLI \cite{tarunesh2021trusting} dataset consists of \textsc{CheckList} templates of (premise, hypothesis) pairs, associated examples, and corresponding entailment, neutral, or contradiction labels. The templates are categorized according to reasoning categories, such as causal, lexical, and syntactic. We report results for evaluation on LoNLI with models trained on subsets of MNLI and optionally the \textit{Prover} generated data in \Cref{tab:results-lonli}. We observe a similar trend, where incorporation of \textit{Prover} generated data helps in improving the overall performance with significant gains across a number of reasoning categories. The improvement in numerical, presupposition, relational, and temporal categories are most prominent. We also find an almost 2\% overall improvement for 5\% MNLI + 5\% Proofs over 10\% MNLI + 0\% Proofs.

\section{Related Work}

We explore generation of multiple-step natural language proofs for the NLI task. For verification and generation, we take inspiration from the automated theorem proving and research in deductive reasoning with Transformers. We also explore how proofs can be further utilized.

\textbf{Categories of Proofs:} The NLP community has introduced several NLU tasks and datasets with accompanying explanations \cite{wiegreffe-marasovic-2021-review}, ranging from highlighted context, natural language based or structured proof trees. Highlighted explanations hardly help in scenarios requiring external knowledge. Human-provided natural language explanations \cite{camburu2018snli} are hard to validate; and structured explanations are hard to scale. For certain closed-world setting \cite{DBLP:conf/acl/TafjordDC21} and symbolic domains, synthetically generated explanations has been popular.  Authors in \cite{DBLP:journals/corr/abs-2112-00114} explore \textsc{NaturalProofs}, where the proof consists of both natural language and mathematical symbols. However, the underlying reasoning task being mathematical, makes the task more well-defined. A large body of work in the ATP community \cite{DBLP:conf/aaai/PaliwalLRBS20,aygun2020learning,agarwal2021analyzing} generates synthetic fine-grained steps as proofs for symbolic domains. Here,  we define constraints over natural language proofs to ease validation and generation.

\textbf{Iterative Proof Generation and Search:} \citet{DBLP:conf/acl/TafjordDC21} utilize synthetically generated end-to-end proof trees as supervision. Authors train a strong encoder-decoder model such as T5 to iteratively generate the next step of a proof, and use search to generate full proofs. 
Similarly, in the Automated Theorem Proving literature, recent work \cite{DBLP:conf/aaai/PaliwalLRBS20,aygun2020learning,agarwal2021analyzing} has shown that Graph Neural Networks and Transformers  can be trained to perform impressively on the theorem-proving task as part of a neuro-symbolic system. Theorem-proving is a multiple-step search-based task, where the neural method learns to simplify an input goal into several sub-goals. This is done recursively using search, until all sub-goals are proven (i.e., simplified to empty goals). Further, %
These proof systems are easier to define  as the number of operations (or theorems or tactics) comes from a known closed set. Thus, it is non-trivial to extend above methods to generate natural language \textit{proof} without full end-to-end supervision. %

\textbf{Proofs to Improve End-task Accuracy:}  \citet{DBLP:conf/wsdm/HeL0ZW21} and \citet{DBLP:conf/acl/KumarT20} has demonstrated the utility of generating proofs (or natural language explanations). In Knowledge-graph based QA context,  \citet{DBLP:conf/wsdm/HeL0ZW21} show how a teacher network trained on additional intermediate hops can be used to enhance the performance of a student network, that is exposed only to the final output supervision. Most importantly, \citet{DBLP:conf/acl/KumarT20} uses label-specific explanations and use them directly to generate the NLI conclusion. While the authors show how explanations are used to generate the conclusion, the authors do not generate fine-grained steps. Authors also compare with a non-recent baseline by \cite{camburu2018snli} on MNLI, instead of SOTA methods such as RoBERTA-large.

We study multi-step reasoning for the NLI task; where connecting a premise and hypothesis may require external knowledge (including free-form fuzzy commonsense rules). The closest to our work is RuleTaker and ProofWriter \cite{DBLP:conf/acl/TafjordDC21}; where authors explore deductive abilities of Transformers over natural language. We differ from their work, as we do not provide the required rules explicitly and do not restrict the natural language input in any form. Our fact retrieval method is inspired from \citet{DBLP:conf/icml/GuuLTPC20,NEURIPS2020_fc84ad56}. \citet{NEURIPS2020_fc84ad56} proposes to combine a non-parametric retriever model with a parametric generator for knowledge-augmented NLP tasks. While the goals are similar, we use a custom retrieval method as we do not have knowledge-augmented sentence composition supervision for learning retrieval and generation in an end-to-end fashion.

\section{Conclusion}
We propose a method to generate knowledge-enriched multiple-step textual proofs (intermediate conclusions) for the NLI task utilizing only next-step supervision. We train a T5 model to generate the next step given a premise-hypothesis pair, and use external fact augmentation, search for more generalized proofs. To ease generation and automatic verification, we introduce constraints over expected proofs, and associated automated metrics. Both human and automated verification show the effectiveness of our proposed method. We also show that our generated proofs can be used to improve NLI task performance using standard data augmentation techniques (on low-data scenarios).

\bibliography{custom}
\bibliographystyle{acl_natbib}

\clearpage
\appendix

\section{Prover}
The multitask T5 generator model(~\Cref{sec:model-prover}) can be trained on additional objective functions for improved performance and better generalization. The objectives are as follows:

\noindent \textbf{Explanation Generation}: The model is trained to generate an explanation $X$ given the premise $P$ and hypothesis $H$. The input is \code{explain: P <sep> H} and the output to be generated is \code{X}. The non-neutral instances of the E-SNLI~\cite{camburu2018snli} dataset is used for this objective.
\vspace{1mm}

\noindent \textbf{Entailment Bank Proof Generation:} The model is also trained to generate one-step proofs given a premise and a conclusion. Suppose conclusion $c$ can be derived from premise sentences $s_1, s_2$. For this instance, the input to the model is \code{proof: $s_1$ <sep> c} and the output to be generated is \code{$s_2$}. Another instance is also created by interchanging $s_1$ and $s_2$ in the input and output. We consider all leaf sentences, intermediate conclusions, and the final hypothesis of the proof trees in the Entailment Bank dataset~\cite{dalvi2021explaining} to create these instances.

These two objectives are more useful for explanation / intermediate step generation for neutral premise-hypothesis pairs.

\section{Constrained Proof Generation}
\label{sec:cpg}
The \textit{explain:} and \textit{proof:} generation setup can be used to directly generate the proof in using the premise $P$ and hypothesis $H$.
We observe that the generator model learns to exploit the proof generation process by generating a single sentence as proof, due to the nature of the training data in E-SNLI and Entailment Bank. We thus evaluate the generated proofs against the gold test set annotations in E-SNLI with a number of text generation evaluation metrics -- BLEU, METEOR, ROUGE, CIDER and semantic similarity (SIM) using \textit{all-mpnet-base-v2}. The results are reported in \Cref{tab:results-constrained}.

\begin{table}[ht!]
\centering
\resizebox{\linewidth}{!}{
	\begin{tabular}{l|cccccc|}
	\toprule
	\textbf{Prefix} & \textbf{BLEU1} & \textbf{METEOR} & \textbf{ROUGE} & \textbf{CIDEr} & \textbf{SIM}\\
	\midrule
	\textit{explain} & 0.4063 & 0.2491 & 0.3898 & 1.6603 & 0.6372 \\
	\textit{proof} & 0.2631 & 0.1537 & 0.2698 & 0.7705 & 0.4720  \\
    \bottomrule
	\end{tabular}
	}
	\caption{Results for constrained proof generation. The generated proofs are matched against the gold annotations in E-SNLI. \textbf{SIM} indicates semantic similarity.}
	\label{tab:results-constrained}
\end{table}

\section{Datasets used}
All datasets used and results reported in this paper are for English language.

\paragraph{LoNLI:}
The dataset contains templates of \textit{(premise, hypothesis)} pairs, associated examples, and corresponding entailment, neutral, or contradiction labels. For instance, one such entailment template is - \textit{premise: {Name} moved from {Country1} to {Country2}; hypothesis: {Name} now lives in {Country2}}. All examples created from this template are labeled as entailment. There are 363 templates in total (166 entailment, 163 contradiction, 34 neutral) each having 1000 examples. The templates are categorized according to reasoning categories, such as lexical, syntactic, boolean, causal, etc. The dataset was inspired by the behavioral testing methodology of NLP systems proposed in C\textsc{heck}L\textsc{ist}~\cite{ribeiro-etal-2020-beyond}.

\section{Automatic Evaluation of Gold Proofs}
We report the automatic evaluation metrics of \textit{Gold Proofs}, \textit{Negated Gold Proofs}, and \textit{Perturbed Gold Proofs} in \Cref{tab:results-multi}. 
For \textit{Negated Gold Proofs}, we take the intermediate steps of the \textit{Gold Proofs} and transform them to a contradictory sentence. We perform this operation using the contradiction generator of the \textit{Prover} model (\Cref{sec:model-prover}). This resultant \textit{proof} is now incorrect, which is accurately represented by the very low correctness scores in \Cref{tab:results-multi}. For \textit{Perturbed Gold Proofs}, we randomly replace 50\% of the words in the intermediate steps with a DistilBERT contextual augmenter~\cite{ma2019nlpaug}. As expected, we observe significant drops across correctness, BLEU4, and Jaccard similarity metrics in \Cref{tab:results-multi}. The drop in BLEU4 and Jaccard similarity is more prominent for \textit{Perturbed Gold Proofs} compared to \textit{Negated Gold Proofs}. This result is intuitive because of the word replacing strategy in \textit{Perturbed Gold Proofs}. We conclude that the proposed automatic evaluation metrics capture the properties and constraints of \textit{proofs} distinctly and are thus suitable for the task.  

\section{Additional Results}
\begin{table*}[h!]
\centering
\resizebox{\linewidth}{!}{
	\begin{tabular}{l|c|c|c|c|cc|cc|cc|cc|c|ccc}
	\toprule
	\multirow{3}{*}{\textbf{Algorithm}} & \multicolumn{4}{c|}{\textbf{Correctness}} & \multicolumn{12}{c}{\textbf{Minimality}}\\
	& \multirow{2}{*}{$P-H$} & \multirow{2}{*}{$P-I_1$} & \multirow{2}{*}{$I_1-I_n$} & \multirow{2}{*}{$I_n-H$} & \multicolumn{2}{c|}{$P-H$} & \multicolumn{2}{c|}{$P-I_1$} & \multicolumn{2}{c|}{$I_1-I_n$} & \multicolumn{2}{c|}{$I_n-H$} & \multirow{2}{*}{$\#$ Steps} & \multicolumn{3}{c}{$\#$ Keywords} \\
	& & & & & B4 & JS & B4 & JS & B4 & JS & B4 & JS & & P & $I_1-I_n$ & H\\
	\midrule
	Gold Proofs & \multirow{3}{*}{95.27} & 94.59 & 98.42 & 94.09 & \multirow{3}{*}{0.0272} & \multirow{3}{*}{66.26} & 0.0981 & 68.34 & 0.2223 & 74.07 & 0.0911 & 65.86 & \multirow{3}{*}{2} & \multirow{3}{*}{6.82} & 2.94 & \multirow{3}{*}{2.54} \\
	Negated Gold Proofs & & 3.94 & 9.45 & 20.08 & & & 0.0189 & 61.61 & 0.0596 & 62.59 & 0.0502 & 59.91 & & & 2.83 & \\
	Perturbed Gold Proofs & & 5.91 & 3.15 & 24.41 & & & 0.0139 & 62.37 & 0.0202 & 58.30 & 0.0147 & 56.80 & & & 3.10 & \\
	\midrule
	T5 w/o Search & 92.28 & 94.06 & 94.63 & 36.25 & 0.0751 & 71.07 & 0.0793 & 67.36 & 0.3070 & 78.82 & 0.1162 & 67.45 & 2 & 7.14 & 2.74 & 3.27 \\
	\midrule
	UC Level Search & \multirow{2}{*}{92.28} & 92.46 & 89.85 & 67.85 & \multirow{2}{*}{0.0751} & \multirow{2}{*}{71.07} & 0.1367 & 70.97 & 0.2443 & 75.94 & 0.0987 & 68.76 & 2 & \multirow{2}{*}{7.14} & 3.25 & \multirow{2}{*}{3.27} \\
	UC Beam Search &  & 91.99 & 82.55 & 72.94 &  &  & 0.1308 & 69.81 & 0.2652 & 75.82 & 0.2240 & 76.09 & 1.71 & & 3.21 & \\
	\midrule 
	Proof Search with Facts & 92.42 & 68.69 & 67.68 & 90.15 & 0.0066 & 65.52 & 0.6261 & 90.10 & 0.3540 & 80.63 & 0.0826 & 71.05 & 2.93 & 7.36 & 4.44 & 2.92 \\
	Proof Search with Facts* & 92.42 & 98.23 & 93.43 & 88.89 & 0.0066 & 65.52 & 0.4457 & 89.83 & 0.2699 & 79.93 & 0.0796 & 70.33 & 2.93 & 7.36 & 5.45 & 2.92 \\
    \bottomrule
	\end{tabular}
	}
	\caption{Additional results for proof search with facts. All the rows except the last row are the same results as reported in Table 3. The last row Proof search with facts* shows the additional results. Proof search with facts* indicate the setting where the fact being composed with a given step is concatenated with that particular step for the calculation of the correctness and minimality metrics. Proof search with facts* results in higher correctness and closer minimality scores relative to the gold proofs.}
	\label{tab:results-multi-appendix}
\end{table*}
\begin{table*}[ht!]
\small
\centering
\resizebox{0.8\linewidth}{!}{
	\begin{tabular}{cc|cccc|ccc}
	\toprule
	\multicolumn{2}{c|}{\textbf{Trained On}} & \multicolumn{4}{c|}{\textbf{MNLI}} & \multicolumn{3}{c}{\textbf{MED}} \\
	\textbf{MNLI \%} & \textbf{Proofs \%} & \textbf{Entailment} & \textbf{Neutral} &\textbf{Contradict} & \textbf{Average} & \textbf{Entailment} & \textbf{Neutral} & \textbf{Average} \\
	\midrule
	1 & 0 & 87.09 & 81.94 & 90.01 & 86.41 & 39.62 & 38.12 & 38.87 \\
	1 & 1 & 89.08 & 81.30 & 90.03 & 86.91 & 45.00 & 33.93 & 39.49 \\
	\midrule 
	2 & 0 & 89.45 & 84.59 & 91.63 & 88.61 & 40.44 & 38.39 & 39.42 \\
	2 & 2 & 89.67 & 84.62 & 91.87 & 88.78 & 46.52 & 39.58 & \textbf{43.07} \\
	\midrule 
	10 & 0 & 91.27 & \textbf{86.12} & \textbf{92.97} & \textbf{90.19} & 43.12 & 37.42 & 40.28 \\
	5 & 5 & 90.18 & 85.59 & 92.56 & 89.50 & 42.45 & \textbf{39.71} & 41.09 \\
	10 & 10 & \textbf{91.37} & 84.93 & 92.32 & 89.63 & \textbf{48.23} & 36.40 & 42.35 \\
    \bottomrule
	\end{tabular}
	}
	\caption{F1 scores for MNLI (val-matched) and MED datasets for the DeBERTa-Large model. 
	Scores are average of 3 runs.}
	\label{tab:results-mnli-med-deberta}
\end{table*}

\begin{table*}[h!]
\small
\centering
\resizebox{0.8\linewidth}{!}{
	\begin{tabular}{l|ccc|ccc}
	\toprule
	\textbf{Data (MNLI \%, Proofs \%)} $\rightarrow$ & 1\%, 0\% & 2\%, 0\% & 1\%, 1\% & 10\%, 0\% & 5\%, 5\% & 10\%, 10\% \\
	\textbf{Category} $\downarrow$ & & & & & & \\
	\midrule 
	Boolean & 47.19 & \textbf{64.66} & 62.51 & \textbf{66.33} & 64.98 & 62.29 \\
    Causal & 96.74 & \textbf{97.23} & 97.13 & \textbf{98.76} & 97.78 & 97.53 \\
    Comparative & 60.12 & 62.64 & \textbf{62.69} & 62.17 & \textbf{70.46} & 63.10 \\
    Conditional & 42.62 & 54.49 & \textbf{56.10} & 40.90 & \textbf{65.53} & 53.41 \\
    Coreference & \textbf{69.13} & 62.50 & 62.31 & \textbf{87.71} & 74.21 & 78.58 \\
    Implicature & 29.70 & 33.23 & \textbf{42.34} & 33.44 & \textbf{39.41} & 35.73 \\
    Lexical & 96.09 & 96.72 & \textbf{97.00} & 93.17 & \textbf{97.19} & 96.80 \\
    Negation & 95.49 & 96.37 & \textbf{99.20} & 95.26 & \textbf{98.21} & 95.49 \\
    Numerical & 50.94 & \textbf{62.46} & 61.78 & \textbf{78.06} & 71.23 & 68.35 \\
    Presupposition & 78.79 & \textbf{90.64} & 88.37 & 95.41 & 95.17 & \textbf{95.45} \\
    Quantifier & 53.81 & \textbf{71.44} & 58.44 & 59.48 & 71.17 & \textbf{73.67} \\
    Relational & 96.50 & \textbf{98.25} & 96.10 & 97.03 & 96.74 & \textbf{99.15} \\
    Spatial & 56.26 & 58.58 & \textbf{62.33} & 51.05 & 56.15 & \textbf{64.47} \\
    Syntactic & 93.86 & 99.77 & \textbf{99.94} & \textbf{100.0} & \textbf{100.0} & \textbf{100.0} \\
    Taxonomic & 66.67 & 69.84 & \textbf{72.60} & \textbf{75.14} & 68.56 & 73.65 \\
    Temporal & 24.70 & \textbf{45.37} & 37.82 & \textbf{55.02} & 52.76 & 52.31 \\
    World & 99.50 & \textbf{99.75} & 99.50 & \textbf{99.50} & \textbf{99.50} & \textbf{99.50} \\
    \midrule
    \textbf{Average} & 68.13 & \textbf{74.35} & 73.89 & 75.79 & \textbf{77.59} & 77.03 \\
    \bottomrule
	\end{tabular}
	}
	\caption{Reasoning category wise F1 scores on the Lo-NLI dataset for the DeBERTa-Large model. Scores are average of 3 runs.}
	\label{tab:results-lonli-deberta}
\end{table*}
\begin{table*}[t!]
\small
\centering
\begingroup
\resizebox{0.94\linewidth}{!}
{\begin{tabular}{L{0.5cm}L{4.8cm}L{3.6cm}L{6.4cm}}
\toprule
\# & \textbf{Premise (P)} & \textbf{Hypothesis (H)} & \textbf{Proof}\\
\midrule
i) & \begin{tabular}{@{}L{4.8cm}} A black-haired man is entertaining a crowd with a hula hoop. \\ \end{tabular}
&  
A man has black hair.
& 
\begin{tabular}{@{}L{6.4cm}}
\textbf{P $\rightarrow$ I1:} A man with black hair is performing.  \\
\textbf{I1 $\rightarrow$  H}
\end{tabular}
\\
\midrule

ii) & \begin{tabular}{@{}L{4.8cm}} A police officer standing next his police motorbike holding a yellow safety jacket. \\ \end{tabular}
&  
A police officer and his motorbike.
& 
\begin{tabular}{@{}L{6.4cm}}
\textbf{P $\rightarrow$ I1:} Police are riding a police motorbike.  \\
\textbf{I1 $\rightarrow$ I2:} A police motorbike is being rode by an officer.  \\
\textbf{I2 $\rightarrow$  H}
\end{tabular}
\\
\midrule

iii) & \begin{tabular}{@{}L{4.8cm}} A person in a green robe sits on a couch with a blanket.  \\ \end{tabular}
&  
A person in a robe sits on a couch.
& 
\begin{tabular}{@{}L{6.4cm}}
\textbf{P $\rightarrow$ I1:} A woman in a green robe sits on a couch.  \\
\textbf{I1 $\rightarrow$ I2:} A girl in a green robe sits on a couch.  \\
\textbf{I2 $\rightarrow$  H}
\end{tabular}
\\
\midrule

iv) & \begin{tabular}{@{}L{4.8cm}} A man and wife stand at the alter as they get married. \\ \end{tabular}
&  
A couple is getting married.
& 
\begin{tabular}{@{}L{6.4cm}}
\textbf{P $\rightarrow$ I1:} A man and a woman are married.  \\
\textbf{I1 $\rightarrow$ I2:} A husband and a wife are married.  \\
\textbf{I2 $\rightarrow$  H}
\end{tabular}
\\
\midrule

v) & \begin{tabular}{@{}L{4.8cm}} A man gets a skateboard up on a big rock. \\ \end{tabular}
&  
The skateboarder is near a rock.
& 
\begin{tabular}{@{}L{6.4cm}}
\textbf{P $\rightarrow$ I1:} A person puts his skateboard on a rock.  \\
\textbf{I1 $\rightarrow$ I2:} Person skateboarding on a rock.  \\
\textbf{I2 $\rightarrow$  H}
\end{tabular}
\\
\midrule

vi) & \begin{tabular}{@{}L{4.8cm}} A professional swimmer spits water out after surfacing while grabbing the hand of someone helping him back to land. \\ \end{tabular}
&  
A person is swimming.
& 
\begin{tabular}{@{}L{6.4cm}}
\textbf{F1:} A hand is part of the arm.  \\
\textbf{F1 $\& $ P $\rightarrow$ Monotone $\rightarrow$ I1} A person is swimming. \\
\textbf{I1 $\rightarrow$  H} \\
\end{tabular}
\\
\midrule

vii) & \begin{tabular}{@{}L{4.8cm}} A female guitarist is playing on stage. \\ \end{tabular}
&  
A woman is playing her instrument.
& 
\begin{tabular}{@{}L{6.4cm}}
\textbf{F1:} A guitar is an instrument.  \\
\textbf{F1 $\& $ P $\rightarrow$ Monotone $\rightarrow$ I1} A woman is playing an instrument. \\
\textbf{I1 $\rightarrow$  H} \\
\end{tabular}
\\
\midrule

viii) & \begin{tabular}{@{}L{4.8cm}} Two middle aged police officers watch over a parking lot at night. \\ \end{tabular}
&  
A couple cops keep an eye on the parking lot.
& 
\begin{tabular}{@{}L{6.4cm}}
\textbf{F1:} An eye is used to see with.  \\
\textbf{F1 $\& $ P $\rightarrow$ Monotone $\rightarrow$ I1} Some cops look at the parking lot. \\
\textbf{I1 $\rightarrow$  H} \\
\end{tabular}
\\
\bottomrule
\end{tabular}
}
\endgroup
\caption{Some examples of generated proofs. Analysis can be found in \Cref{sec:appendix-analysis}.}
\label{tab:examples-appendix}
\end{table*}

We also show additional results for Proof Search with Facts in \cref{tab:results-multi-appendix}. We mentioned earlier in \Cref{sec:auto-eval} that the recursive composition of facts result in reduced correctness scores, as pairs are not always entailed  because of the introduction of new knowledge. This is demonstrated in the Proof Search with Facts row in \cref{tab:results-multi-appendix}. We evaluate another method Proof search with facts*, where the fact being composed with a given step is concatenated with that particular step for the calculation of the correctness and minimality metrics. This strategy results in higher correctness and closer minimality scores relative to the gold proofs.

We show some additional results to show the usefulness of the \textit{Prover} for various NLI tasks. This is an extension of the results reported in \cref{sec:usefulness}. As described earlier, we use the \textit{Prover} generated data as additional labeled data for supervised learning in NLI tasks. Here, we train a DeBERTa-Large model~\cite{he2021debertav3} with i) MNLI-only data or ii) a mix of MNLI data and \textit{Prover} generated data. The results are reported for MNLI and MED datasets in \cref{tab:results-mnli-med-deberta}, and for Lo-NLI dataset in \cref{tab:results-lonli-deberta}. 

For the DeBERTa-Large model, we found similar conclusions that we made earlier for the RoBERTa-Large model in \cref{sec:usefulness}. The incorporation of \textit{Prover} generated data helps in improving performance across most settings in the three datasets. In the very low data regime (1\% MNLI + 0\% Proofs \textit{vs.} 1\% MNLI + 1\% Proofs or 2\% MNLI + 0\% Proofs \textit{vs.} 2\% MNLI + 2\% Proofs), the additional data from \textit{Prover} helps in improving average performance in the MED dataset. 

The incorporation of \textit{Prover} generated data also helps in improving the overall performance with significant gains across a number of reasoning categories in the Lo-NLI dataset. We find an 1.8\% overall improvement for 5\% MNLI + 5\% Proofs over 10\% MNLI + 0\% Proofs.

\section{Analysis of Generated Proofs}
\label{sec:appendix-analysis}
We manually analyze proofs from both \textit{unconstrainted search}  and \textit{proof search with facts}. Some interesting example generated proofs are in \Cref{tab:examples-appendix}. In example i) and ii), the premise and the hypothesis are lexical/syntactic paraphrases. Hence, no intermediate steps are required as a proof. In example iii), undue specialization or hallucinations are introduced in the intermediate steps, as the word \textit{person} is changed to a \textit{woman} and \textit{girl}. In example iv), additional knowledge beyond the ones retrieved are required to proof the instance. In this case, the knowledge \textit{a man and his wife are called a couple}, or \textit{a woman and his husband are called a couple} would be required. Example v) requires spatial reasoning about physical objects \textit{skateboard} and \textit{rock}. In example vi), an unrelated fact is retrieved which is not useful for the proof. Example vii) and viii) are instances of correct proofs. We also observe that fact composition provides consistently superior results, especially when ontological knowledge (IsA, HasA, is-part-of relations) are involved. However, as both the search procedures does not involve any learning, in some case intermediate steps are generated that moves the reasoning away from the hypothesis -- for example ``men'' converted to ''people', then converted back to ``male'' in the following example; ``a \textit{man} in a black shirt and brown pants is jumping in the air performing a karate kick towards a \textit{man} in a red shirt and black hat'' $\rightarrow$
``two \textit{men} are performing karate'' $\rightarrow$
``two \textit{people} are doing martial arts''
$\rightarrow$
``two \textit{males} are involved in martial arts.''

\section{Instructions for Human Verification}
We performed human verification for proofs with two intermediate steps. The following instructions were given as it is to the human annotators.
\subsection{Definition of Entailment}
Entailment is a directional relation between two sentences - S1 and S2. The relation holds whenever the truth of the second sentence S2 follows from the first sentence S1. 
In other words, if a human reading S1 infers that S2 is true, then (S1, S2) is an entailment pair. Note that, (S1, S2) being an entailment pair does not necessarily mean that the reverse pair (S2, S1) is an entailment pair. 
Some examples are given below:
\begin{itemize}
\item S1: A football game with multiple males playing.

S2: Some men are playing a sport.

Label: (S1, S2) is an entailment pair.
\end{itemize}

However, (S2, S1) is not an entailment pair because playing a sport does not necessarily mean playing football. More examples:
\begin{itemize}
\item S1: A woman is walking outside.

S2: A person outdoors.

Label: (S1, S2) $\rightarrow$ Entailment.

\item S1: An older and younger man smiling.

S2: Two men are smiling and laughing at the cats playing on the floor.

Label: (S1, S2) $\rightarrow$ Not Entailment.
\end{itemize}

If S1 and S2 are the same sentences or very similar sentences with minimal difference in tokens then consider that as entailment.
\begin{itemize}
\item S1: An older and younger man smiling.

S2: An older and younger man smiles.

Label: (S1, S2) $\rightarrow$ Entailment.
\end{itemize}

\subsection{Definition of Uniqueness}
S1 and S2 are two given sentences. If S1 and S2 are very similar in the token space then consider (S1, S2) as not unique. If they are not very similar then they are unique.
This is a symmetrical relation. The uniqueness of (S1, S2) is the same as (S2, S1).

\begin{itemize}
\item S1: A football game with multiple males playing.

S2: Some men are playing a sport.

Label: Unique.
	
\item S1: An older and younger man smiling.

S2: An older and younger man smiles.

Label: Not unique. 

\item S1: A man is playing a guitar.

S2: A man is playing a musical instrument.

Label: Unique. 

\item S1: A man is playing a guitar.

S2: A guitar is being played by a man.

Label: Not Unique. 
\end{itemize}

Please note the difference between the last two examples carefully. The usage of the word musical instrument makes the third pair unique.

\subsection{Instructions}
Each instance has four elements: i) Premise, ii) Step 1, iii) Step 2, and iv) Hypothesis. We are aiming to find a \textit{proof} of the hypothesis from the premise. In our case, Step 1 and Step 2 constitute the proof that lets us go from the premise to the hypothesis. In other words, Step 1 and Step 2 provide the gradual transition between the premise and the hypothesis. Considering this, we have 3 sentence pairs from each proof:
\begin{itemize}
\item X1 : (Premise, Step 1)

X2 : (Step 1, Step 2)

X3 : (Step 2, Hypothesis)
\end{itemize}

Each instance has to be scored in two aspects using the three above pairs:

\begin{enumerate}
    \item Correctness: Give a score among [3, 2, 1, 0]. This is assigned by checking the entailment label of the three pairs.
    \begin{itemize}
        \item X1, X2, X3 are all entailment: Score is 3
        \item Only X1, X2 are entailment: Score is 2
        \item Only X1 is entailment: Score is 1
        \item All other cases: Score is 0
    \end{itemize}
    Note that the priority of the pairs are: X1 > X2 > X3. If X2, X3 are both entailment but X1 is not then we will give a score of 0.
    
    \item Minimality: Give a score among [1, 0].
    \begin{itemize}
        \item If X1, X2, X3 are all unique pairs: Score is 1
        \item If any pair is not unique: Score is 0 
    \end{itemize}
\end{enumerate}

\section{Experimental Setup}
We use beam search to generate outputs from the T5-Large models. A beam length of 10 is used. The T5-Large models were trained with the Adafactor optimizer~\cite{shazeer2018adafactor} with a learning rate of 5e-6. 
We retrieve top 8 facts (according to cosine similarity) from the retriever model for composition.

\section{Computational Resources}
We use a single Quadro RTX 8000 GPU for our experiments. We train the \textit{Prover} T5 model and composition T5 model for 15 and 5 hours in this GPU.
The T5-Large and RoBERTa-Large models have 770M and 355M parameters, respectively.

\end{document}